\documentclass[11pt]{article}

\usepackage[final]{acl}

\usepackage{times}
\usepackage{latexsym}

\usepackage[T1]{fontenc}
\usepackage[utf8]{inputenc}

\usepackage{microtype}

\usepackage{inconsolata}

\usepackage{graphicx}

\usepackage{amsmath}
\usepackage{amssymb}
\usepackage{booktabs}
\usepackage{listings}
\usepackage[normalem]{ulem}
\title{Synthetic Semantic Supervision for Contrastive Code Representation Learning in Small Transformers: An Empirical Study}

\author{Kenneth Paulsen \\
  University of Luxembourg (SnT) \\
  \texttt{kenneth.paulsen@uni.lu} \\\And
  Florian Tambon \\
  University of Luxembourg (SnT) \\
  \texttt{florian.tambon@uni.lu} \\ \AND
  Mike Papadakis \\
  University of Luxembourg (SnT) \\
  \texttt{michail.papadakis@uni.lu} \\\And
  Shin Yoo \\
  School of Computing, KAIST \\
  \texttt{shin.yoo@kaist.ac.kr}
}

\begin{document}
\maketitle
\begin{abstract}
General-purpose code embeddings power tools for code search, classification, and retrieval. Compact transformer encoders for code typically rely on either human-written docstrings (labor-intensive and inconsistent) or mined structural signals such as execution traces (setting-specific and costly to collect). We empirically study an alternative: contrastive pretraining of small encoders with synthetically generated natural-language descriptions emphasizing code functionality and intent, paired with code in a dual-encoder framework at training and discarded at inference. We benchmark this approach against pretraining-based baselines, generalist LLMs, and embedding-specific models on eight retrieval, classification, and generation tasks across C, C++, and Java. Synthetic semantic supervision yields statistically significant gains over pretraining baselines of the same inference-time size on five of eight tasks, with parity on two more; once fine-tuned, it matches or exceeds zero-shot models two orders of magnitude larger on classification, and it stays on par with execution-aware supervision at matched pretraining data, suggesting a scalable, effective alternative to existing code-representation paradigms.
\end{abstract}

\section{Introduction}\label{sec:introduction}

\begin{figure}[t!]
  \centering
  \includegraphics[width=0.85\columnwidth]{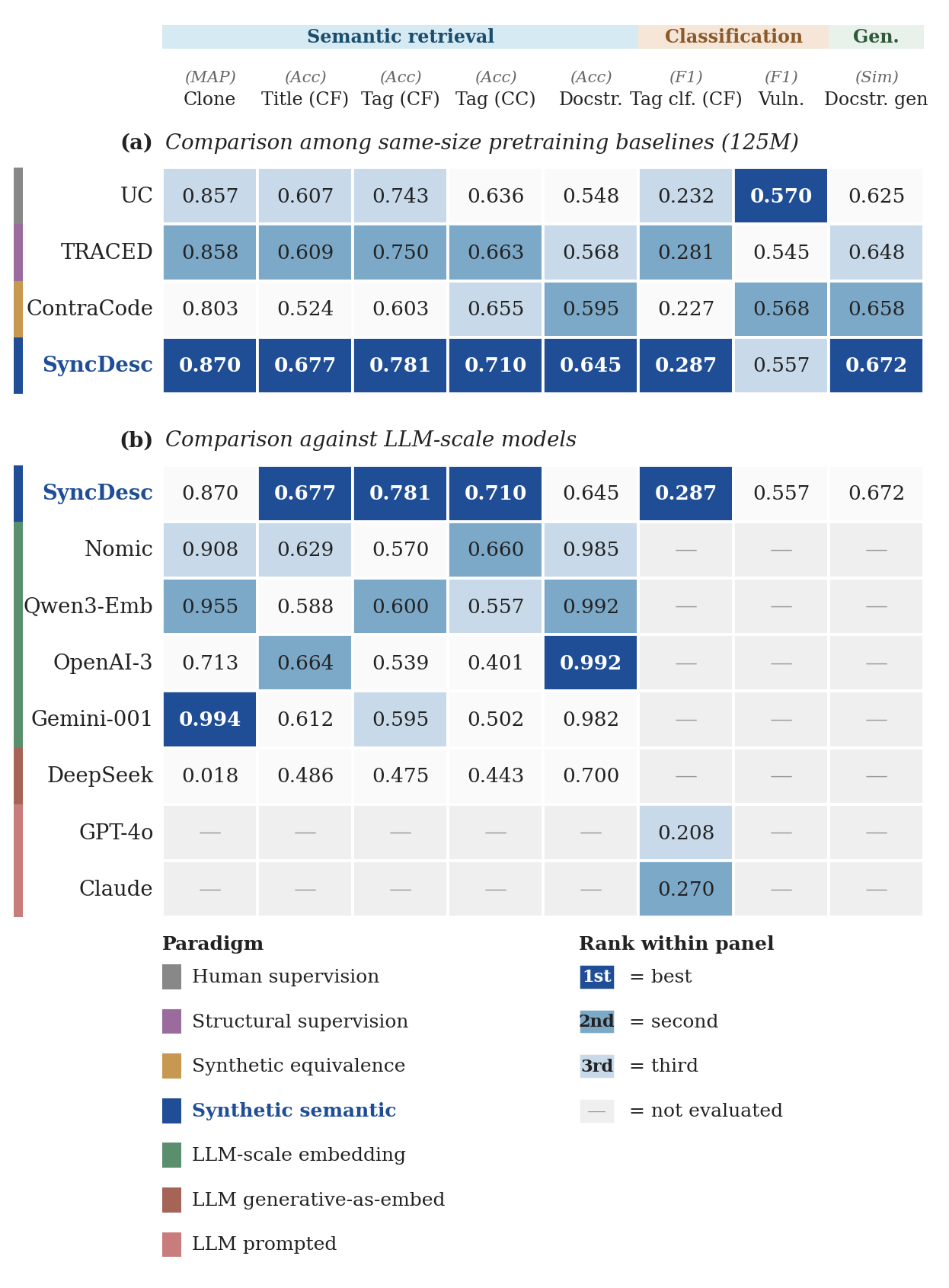}
  \caption{Comparison across eight downstream tasks. Methods grouped by paradigm (left band). Cell color indicates rank within each panel and task column: 1st (dark), 2nd (medium), 3rd (light); cells in columns with a single evaluated method are unshaded; em-dashes indicate the method was not evaluated. \textbf{(a)} Against pretraining baselines of the same inference-time size, SyncDesc improves significantly on five of eight tasks and is at statistical parity on two more (paired bootstrap tests in Table~\ref{tab:significance}); ContraCode's ranking carries a cross-language transfer penalty (JavaScript pretraining; Section~\ref{sec:results}). \textbf{(b)} SyncDesc, fine-tuned per task, remains competitive against zero-shot LLM-scale models on algorithmic retrieval and classification. SyncDesc is best on single-source / dual-source variants (per-variant breakdown in Appendix~\ref{app:detailed_comparison}). Full results in Appendix~\ref{app:full_results}.}
  \label{fig:hero}
\end{figure}
%Modern software systems are built using increasingly large and heterogeneous code bases. Developers routinely work across multiple languages, repositories, and abstraction layers, and core activities such as navigating unfamiliar projects, retrieving relevant examples, or understanding legacy implementations can be time-consuming and cognitively demanding~\cite{xia2017measuring, kohl2020multitasking}. In response to these challenges, 
In recent years, 
a growing number of automated tools for code search, documentation, analysis, and coding assistance has been proposed~\cite{Copilot, guo2022unixcoder, huang2020codedescriptionrep}. These tools rely on effective code representations: abstractions that translate human-written source code into formats suitable for automated processing.\footnote{Code, data, the generated descriptions, and the generation logs are available at \url{https://zenodo.org/records/22130652}.}

A common approach is to use learned embeddings derived from token sequences or structured program elements~\cite{xie2023surveycodesearch}. While Large Language Models (LLMs) such as GPT-4o mini~\cite{openai_gpt4o_mini_2024} can generate such embeddings, and specialized embedding LLMs such as Nomic~\cite{nussbaum2025nomic} can do so as well, their use incurs high computational and financial costs (e.g., API calls, infrastructure) and scales poorly for many practical applications. This holds in particular for \emph{embedding workloads} (code search, retrieval, and clone detection over large candidate pools), where ranking $N$ candidates requires $N$ precomputed, mutually comparable vectors rather than $N$ prompts per query, and per-query API calls add latency, rate limits, and code egress that private codebases often prohibit. Prompted LLMs retain the complementary advantage of zero-shot generalization; the deployment class we target is the former. In contrast, smaller, specialized transformer models have emerged as a cost-effective alternative, achieving strong performance on targeted coding tasks through fine-tuning~\cite{zeng2022extensive, 10.1145/3725732, wang2025comprehensive}.

Prior work on training code embedding models generally falls into two paradigms. The first, human supervision, uses manually written specifications (e.g., docstrings) as supervision signals to capture program semantics, as in UnixCoder ~\cite{guo2022unixcoder} or CodeBERT~\cite{feng2020codebert}. While effective, this approach is labor-intensive and struggles to produce representations that generalize across tasks, languages, or evaluation settings~\cite{li2023plm_overinterpretation}. Moreover, existing annotations often rely on surface-level lexical or syntactic cues, providing only indirect access to semantic intent~\cite{panthaplackel2020associating}, and may contain inconsistencies or errors~\cite{siddiq2024fault}. The second paradigm, structural supervision, leverages dynamic or control-flow information mined directly from code (e.g., traces, instrumentation, or program analysis), as in TRACED~\cite{ding2024traced}. While more automated, these approaches can be expensive to collect and must be tailored to specific languages or settings. Recent work has also explored synthetic supervision for code. For example, ContraCode~\cite{jain2021contrastive} proposed generating synthetic equivalent code to pre-train a neural model, while the Qwen2.5-coder white paper~\cite{hui2024qwen2} demonstrated the use of synthetic data for pretraining code models. These efforts focus primarily on synthetic equivalent code, which limits the diversity of tasks (i.e., they require equivalence), or they train large models, raising the question of the benefit compared to the increase in size. Finally, they focus mainly on general code generation tasks rather than representation learning for downstream understanding tasks.

Despite a wealth of approaches, evaluations and direct comparisons of these approaches remain scarce when it comes to downstream code-oriented tasks. Moreover, while synthetic datasets have proven useful for training LLMs, they remain underexplored for smaller code-oriented transformers. Starting from this observation, we ask: how do existing code-embedding methods fare, and could a simple, scalable approach (using synthetic semantic descriptions and contrastive learning) produce high-quality code embeddings that rival or exceed existing methods? In this paper, we empirically evaluate the following hypothesis: contrastive pretraining with synthetic semantic descriptions can yield code embeddings that (1) outperform traditional pretrained baselines, (2) match or exceed the performance of larger models on downstream tasks, and (3) generalize across languages and tasks, all while being computationally efficient.

To test these hypotheses, we select several baseline approaches from the literature covering a broad spectrum of methods, namely pretraining-based methods (TRACED~\cite{ding2024traced}, ContraCode~\cite{jain2021contrastive}) applied to a small baseline transformer (UnixCoder~\cite{guo2022unixcoder}), large generalist models (GPT-4o~\cite{openai_gpt4o_systemcard_2024}), and large generalist embedding models (Nomic~\cite{nussbaum2025nomic}). We further propose a simple approach, \textit{SyncDesc}, relying on synthetically generated semantic descriptions at scale, paired with code snippets to form aligned supervision for a dual-encoder contrastive objective~\cite{radford2021clip}, exploiting research highlighting their individual effectiveness~\cite{hui2024qwen2, guo2022unixcoder}. SyncDesc is best understood as lightweight knowledge distillation: it transfers the semantic-extraction ability of a large teacher model into a compact encoder through generated descriptions, rather than introducing a new mechanism. In this respect, none of the compared pretraining paradigms is information-free: TRACED injects privileged execution information, ContraCode injects compiler-verified equivalence, and SyncDesc injects teacher-model semantics; the empirical question this paper answers is which external signal is most effective per unit cost at a fixed inference-time size. Notably, the fine-tuned 125M student exceeds its own teacher on the one task where the teacher is evaluated (Section~\ref{sec:results}). We further conduct an ablation study that isolates the two ingredients of SyncDesc (the synthetic semantic descriptions and the contrastive objective), comparing against variants supervised with human-written docstrings and trained non-contrastively, to quantify how much each contributes to the observed gains.

Our results show that contrastively pretrained models with synthetic semantic descriptions improve significantly over pretrained baselines of the same inference-time size on five of eight tasks, with parity on two more and parity-to-below on vulnerability detection (Table~\ref{tab:significance}). On several tasks, SyncDesc, after task-specific fine-tuning that the zero-shot LLMs lack, matches or outperforms larger language models, despite using significantly smaller models. An ablation study confirms that both synthetic descriptions and contrastive learning are critical to these gains, outperforming embeddings supervised by human docstrings or trained non-contrastively. These findings suggest that synthetic semantic supervision, when combined with contrastive learning, offers a scalable, effective, and simple alternative to existing paradigms for code representation learning.

\section{Experimental Setup}
\label{sec:experimental_setup}

%In this section, we detail the main experimental setup used in our experiments throughout the paper.

\subsection{Approaches compared}

In our experiments we consider two main families of approaches. First, pre-trained transformers using different pre-training objectives and tasks. For all, we use as a baseline model a \textbf{UnixCoder}~\cite{guo2022unixcoder} (125M parameters). This model was pre-trained with standard objectives (e.g., masked language modeling / denoising) using human-written docstrings. The first pre-training method is \textbf{TRACED}~\cite{ding2024traced}, an execution-aware baseline trained on CodeNet traces. The second is \textbf{ContraCode}~\cite{jain2021contrastive}, a pre-training method using synthetic code via equivalent functions. Finally, we add our simple baseline, named in the following \textbf{SyncDesc}, relying on contrastive synthetic docstrings, which are detailed in the next paragraph. The second family of approaches is represented by LLMs, either \textbf{generalist closed-source LLMs} (GPT-3.5-Turbo~\cite{openai_gpt35_turbo_2023}, GPT-4o-mini~\cite{openai_gpt4o_mini_2024}, GPT-4o~\cite{openai_gpt4o_systemcard_2024}, and Claude~3 variants~\cite{anthropic_claude3_2024}) used as-is, or \textbf{embedding LLMs}, i.e. LLMs specifically fine-tuned for embedding generation (Nomic Embed Code (7B)~\cite{nussbaum2025nomic}, Qwen3-Embedding-0.6B~\cite{zhang2025qwen3embedding}, OpenAI text-embedding-3-large~\cite{openai2024embeddings}, and Gemini Embedding 001~\cite{lee2025gemini}). As a further reference point, we also evaluate a generative code model used directly as an embedder without any embedding-specific training (DeepSeek Coder (6.7B)~\cite{guo2024deepseekcoder}), whose pooled hidden representations we use as-is.

\paragraph{\textbf{Proposed baseline}}

The proposed baseline SyncDesc introduces a simple three-stage pipeline for learning code representations using contrastive learning. First, synthetic natural-language descriptions are generated for each code snippet using a large language model (in our case GPT-4o), ensuring semantic abstraction and avoiding lexical overlap with the source code. Generating synthetic documentation offers several advantages over reusing existing docstrings. Many benchmarks lack proper or consistent documentation~\cite{siddiq2024fault}, and natural docstrings are often absent, underdeveloped, or weakly aligned with code semantics. Moreover, while using LLMs can introduce errors due to hallucinations, recent studies showed that large language models can generate descriptions that are comparable to, or more semantically consistent than, human-written documentation in some settings~\cite{dvivedi2024comparative}. These code–description pairs are then used to pretrain a dual-encoder architecture with a code encoder (UnixCoder) and a description encoder (a DeBERTa model~\cite{he2020deberta}) with a symmetric contrastive objective, where representations are aligned in a shared embedding space. To improve training, group-aware negative masking excludes pairs from the same problem group from acting as negatives. Finally, after pretraining, only the code encoder is retained and reused for downstream tasks (such as retrieval or fine-tuning) where its fixed-dimensional embeddings serve as input.

% \begin{table*}[t]
% \centering
% % \tiny
% \caption{Model characteristics and pretraining configurations.}
% \label{tab:models}
% \begin{tabular}{lccc}
% \toprule
% Model & Params & Pretrain Data & Loss \\
% \midrule
% CodeBERT & 125M & GitHub (multi) & MLM+RTD \\
% UnixCoder & 125M & GitHub (multi) & Span+Gen \\
% TRACED & 125M & CodeNet Traces & Multi-task \\
% CB,1S & 125M & Java corpus or CodeNet (C) & Contrastive \\
% UC,1S & 125M & Java corpus or CodeNet (C) & Contrastive \\\
% UC,2S & 125M & Java+C & Contrastive \\
% Nomic Embed Code & 7B & GitHub-scale & Contrastive \\
% DeepSeek-Coder & 6.7B & Web + code & Autoregressive \\
% GPT / Claude variants & undisclosed & Web-scale & Autoregressive \\
% \bottomrule
% \end{tabular}
% \end{table*}

\subsection{Datasets and Tasks}\label{sec:experimental_setup-datasets}

\paragraph{\textbf{Pretraining Datasets.}}

All contrastive models are pretrained using their dedicated process: (i) For TRACED, we use their default C corpus derived from CodeNet~\cite{puri2021codenet}, constructed by following the preprocessing pipeline of TRACED and retaining only \emph{unique static code implementations}. Specifically, multiple execution traces corresponding to the same source code (generated under different inputs) are collapsed into a single code instance, and all dynamic traces are discarded. This results in approximately 18k unique C code samples (24,936 samples; 18,266 unique implementations) used for pretraining (equivalently, the $\sim$121k execution traces reported by~\cite{ding2024traced} collapse to $\sim$18k once traces generated from the same source program under different inputs are merged and dynamic information is discarded, since our setup operates at the level of static code rather than traces), with a disjoint evaluation subset of approximately 1k samples; (ii) for ContraCode, we applied their self-supervised contrastive pretraining pipeline (generating semantically-equivalent program variants through compiler-based source-to-source transformations and training with their instance-discrimination, or MoCo, objective) on the same UnixCoder backbone, using the CodeSearchNet JavaScript corpus ($\sim$1.8M methods) released with their work; this matches the architecture across methods, but the corpus remains JavaScript because ContraCode's augmentation stack is JS-specific and does not port to Java/C without new tooling; the resulting language mismatch is a confound we account for in Section~\ref{sec:results}, while data volume ($\sim$1.8M vs.\ $\sim$120k, roughly $15\times$) favors ContraCode; (iii) we use two pre-training datasets for two separate versions of SyncDesc. The first (SyncDesc-1S) only uses a Java corpus aggregating FunCom~\cite{bansal2023funcom}, DeepCom~\cite{hu2020deepcom}, and CONCODE~\cite{iyer-etal-2018-concode} method-level examples, yielding a pool of $\sim$200k generated code--description pairs, of which 99,991 Java samples (97,252 unique implementations) form the pretraining set after filtering and generation failures (details in Appendix~\ref{app:generation}); The second (SyncDesc-2S) is a mix of CodeNet C corpora (same as TRACED) and Java samples in equal proportion. Note that this Java corpus is mined from GitHub repositories, not competitive programming; only the C corpus derives from CodeNet.

% \begin{table*}[t]
% \centering
% % \tiny
% \caption{Summary of datasets and evaluation metrics. Subsampling (*) is performed via uniform random sampling without replacement.}
% \label{tab:datasets}
% \resizebox{\textwidth}{!}{
% \begin{tabular}{lcccccccc}
% \toprule
% Task & Task Type & Dataset & Lang & Train & Valid & Test & Metric \\ %& Used in RQs \\
% \midrule
% Clone Retrieval & Doc. Retrieval & CodeXGLUE-POJ104 & C/C++ & 32K & 8K & 12K & MAP@499 \\ %& RQ1, RQ2 \\
% Vulnerability Detection & Binary Classif. & CodeXGLUE (Devign) & C & 21.9K & 2.7K & 2.7K & Acc, F1 \\ %& RQ1 \\
% Docstring Matching & Doc. Retrieval & CodeSearchNet & Java & 100K* & 14.1K & 50K* & Rank Acc, Sep.\\ % & RQ1, RQ2 \\
% Tags Matching & Doc. Retrieval & CodeChef (CC) & Java & 13.8K & – & 1.9K & Rank Acc, Sep. \\ %& RQ1, RQ2 \\
% Tags + Title Matching & Doc. Retrieval & CodeForces (CF) & C/C++ & 100K* & 10K & 10K & Rank Acc, Sep. \\ %& RQ1, RQ2 \\
% Tags Classification & Mult. Classif. & CodeForces (CF) & C/C++ & 100K* & 10K & 5K & F1 \\ %& RQ1, RQ2 \\
% Description Generation & Generation & CodeSearchNet & Java & 100K* & 14.1K & 50K* & Sim.\\ % & RQ1 \\
% Clone Detection & Binary Classif. & BigCloneBench & Java & 30K* & 1K* & 10K* & Acc \\ %& RQ3 \\
% Tags Classification & Mult. Classif. & CodeChef (CC) & Java & 13.8K & – & 1.9K & F1 \\ %& RQ3 \\
% \bottomrule
% \end{tabular}}
% \end{table*}

\paragraph{\textbf{Downstream Tasks Datasets.}}
We evaluate on \textbf{eight} downstream tasks spanning retrieval, classification, and generation that were used previously in the literature~\cite{ding2024traced, alon2018code2seq}. For small transformer models (i.e., pre-trained with TRACED, ContraCode, and SyncDesc), we fine-tune them using available train datasets. LLMs are used as-is without further fine-tuning. We describe the tasks in the following. \textbf{Clone Detection} -- \textit{Retrieval setting} (CodeXGLUE-POJ104; \cite{mou2016poj104}): Given a code query and a collection of candidate programs, the task retrieves semantically equivalent implementations (same problem) from the candidate set. \textbf{Clone Detection} -- \textit{Binary classification setting}  (BigCloneBench\cite{Svajlenko2015EvaluatingCD}): Given a pair of code snippets, the task predicts whether the two programs are semantic clones. This dataset also contains human-written docstrings. \textbf{Vulnerability Detection} (CodeXGLUE -- Defect Detection; \cite{zhou2019devign}): Given a source code function, the task predicts whether it is vulnerable (binary classification). \textbf{Docstring Matching} (CodeSearchNet Java; \cite{husain2019codesearchnet}): Given a Java function, the task retrieves the corresponding docstring from a candidate pool. \textbf{Tags Matching} (CodeChef Java; \cite{idialu2024whodunit}): Given a Java code snippet, the task retrieves the corresponding algorithmic tag from a candidate pool (\textbf{98} tag classes). \textbf{Tags and Title Matching} (CodeForces; \cite{penedo2025codeforces}): Given a competitive programming code snippet, the task retrieves the corresponding tag(s) and title from candidate pools. \textbf{Tags Classification} -- \textit{CodeForces} (\cite{penedo2025codeforces}): Given a competitive programming code snippet, the task predicts the tag among the ten most frequent CodeForces tags (multi-class classification). \textbf{Tags Classification} -- \textit{CodeChef Java}~\cite{idialu2024whodunit}: Given a Java code snippet, the task predicts the tag among the ten most frequent CodeChef algorithmic tags (multi-class classification). This dataset also contains human-written docstrings. \textbf{Description Generation} (CodeSearchNet Java; \cite{husain2019codesearchnet}): Given a Java function, the task generates a natural language description of the code. Four of the eight tasks thus evaluate on GitHub-mined or real-project code (BigCloneBench, Devign, and the two CodeSearchNet-Java tasks).

\paragraph{\textbf{Evaluation Metrics.}} For each task we follow the evaluation protocol of the original benchmark or prior work: ranking metrics (MAP@K and ranking accuracy) and similarity separation for retrieval~\cite{mou2016poj104,husain2019codesearchnet,penedo2025codeforces,idialu2024whodunit}, accuracy and F1 for classification~\cite{zhou2019devign,Svajlenko2015EvaluatingCD}, and embedding-based semantic similarity against reference docstrings for generation~\cite{husain2019codesearchnet}. Similarity Separation, used as an additional analysis metric, measures the margin between cosine similarity of the query to its positive match versus its negatives; larger values indicate better separation between matched and non-matched pairs and complement ranking accuracy by capturing embedding geometry rather than rank order~\cite{liu2024contrastive}. 

\subsection{Hyperparameters and Configuration}

\paragraph{\textbf{Description Generation.}}
For our baseline, descriptions are generated with GPT-4o using temperature 0.7, top-$p$ 1.0, and a maximum generation length of 70 tokens. The prompt explicitly instructs the model to produce descriptions of at most 50 tokens, while the higher generation limit serves as a safety margin to prevent truncation. We generated a synthetic docstring for each of the examples in our pretraining datasets. The exact prompt, per-corpus filtering and generation-failure statistics, the full one-time generation cost, and an infrastructure comparison with execution-trace collection are given in Appendix~\ref{app:generation}; the decontamination audit between pretraining corpora and all test sets is given in Appendix~\ref{app:decontamination}.

\paragraph{\textbf{Contrastive Pretraining.}}
Pretraining in all cases uses AdamW with learning rate $2\times10^{-5}$, 10\% warmup, mixed precision, and sequence lengths of 512 (code) and 125 (text). SyncDesc-1S is trained for two epochs; SyncDesc-2S models for five epochs. Per-GPU batch size is 3–4 with gradient accumulation of 4.

\paragraph{\textbf{Downstream Fine-tuning.}}
Fine-tuning uses AdamW with learning rate $2\times10^{-5}$ and 10\% warmup for all methods requiring pre-training (TRACED, ContraCode and SyncDesc). Each experiment is run with three random seeds (42, 123, 456), and results are averaged. 

\textit{Data} For fine-tuning the models on the downstream tasks, we used the available train set, except for tasks with over 100K samples. In that case, we sampled uniformly at random and in a way that guarantees a representative sample at a 95\% confidence interval. When evaluating, we used only the representative sampled test dataset.

\textit{Loss} We use task-appropriate objectives as a loss function:
(i) \emph{contrastive retrieval tasks} (Clone Detection, Docstring Matching, Tags/Title Matching) are fine-tuned with a contrastive objective implemented as cross-entropy over the in-batch similarity matrix between the two task-specific embedding sets (e.g., code--code for clone detection, code--docstring for docstring matching, and code--label text for tags/title matching);
(ii) \emph{classification tasks} (Vulnerability Detection, Tags Classification) are fine-tuned with standard cross-entropy loss; and
(iii) \emph{generation} (Description Generation) uses token-level cross-entropy with teacher forcing.

\textit{Epochs.} Unless stated otherwise, all downstream tasks are fine-tuned for \textbf{1 epoch}. Vulnerability detection is fine-tuned for \textbf{5 epochs}, selecting the best checkpoint by validation dataset performance, following the same protocol used in TRACED.

% \textit{Which models are fine-tuned.} All 125M-scale encoders (CodeBERT, UnixCoder, TRACED, and our pretrained variants) are fine-tuned per downstream task. Nomic Embed Code and DeepSeeker Code are evaluated \emph{without} task-specific fine-tuning (embedding-only retrieval). Closed-source LLMs are evaluated \emph{zero-shot} via prompting only on classification tasks where outputs are directly comparable.

\textit{Generation protocol.} For Description Generation, we keep the code encoder \emph{frozen} and train only the decoder components used for generation, to assess whether pretrained representations support generation without adapting the encoder.

\paragraph{\textbf{Hardware.}}
All experiments run on 4$\times$NVIDIA A100 GPUs (40GB). Seeds are fixed and configurations are logged. 
% All results and scripts are available in our replication package~\cite{REPL_PACKAGE_CITATION_TODO}. 

\section{Results}
\label{sec:results}

% \Florian{This can be removed}
% This section reports empirical results addressing the three research questions defined in Section~\ref{sec:experimental_setup}.
% All values are averaged over three random seeds (42, 123, 456). For language-specific tasks, cells marked ``--'' indicate that the evaluation is not applicable.

% ======================================================
% RQ1: MAIN CONSOLIDATED RESULTS (MID-SCALE MODELS)
% ======================================================
\subsection{Comparison to transformer baselines}

% Table~\ref{tab:rq1_main} reports results across all downstream tasks described in
% Section~\ref{sec:experimental_setup-datasets}.

\textbf{Models pretrained with synthetic semantic descriptions improve significantly over the standard transformer baselines and the execution-aware model on five of eight tasks, with statistical parity on two more.} Figure~\ref{fig:hero} reports all downstream tasks. Gains are largest on semantically oriented retrieval and matching: clone retrieval (MAP 0.870 vs.\ 0.858), title matching (0.677 vs.\ 0.609), tag matching (0.781/0.710 vs.\ 0.750/0.663 on CodeForces/CodeChef), and docstring matching (0.645 vs.\ 0.595); all four are statistically significant under paired bootstrap tests (Table~\ref{tab:significance}). Tag classification (F1 0.287 vs.\ 0.281) and description generation (0.672 vs.\ 0.658) are better on average but not significantly so, and~vulnerability detection stays within a narrow band across all models (F1 0.557 vs.\ 0.570, n.s.; above TRACED at matched pretraining data, 0.557 vs.\ 0.545), consistent with its reliance on fine-grained control-flow rather than semantic supervision. ContraCode, which trains on synthetically generated equivalent code variants, scores at or below the standard transformer baselines (UC) on most tasks; part of this gap is a corpus confound rather than the supervision paradigm: ContraCode is pretrained on JavaScript (its compiler-based augmentations are JS-specific; Section~\ref{sec:experimental_setup-datasets}) and evaluated cross-language, although architecture is matched (re-executed on the same UnixCoder backbone; the original release used a smaller six-layer transformer) and data volume favors it by roughly $15\times$. Its language-boundedness is a property of equivalence-based supervision, noted descriptively rather than as a performance conclusion. The dual-source variant helps most when the single-source corpus is narrow (C/CodeNet); on Java tasks where the single-source corpus already aggregates three datasets, dual-source provides no consistent gain. A full per-task breakdown and the single- versus dual-source analysis are provided in Appendix~\ref{app:detailed_comparison}.

\paragraph{\textbf{Confound-free comparisons.}} For all C/C++ tasks, SyncDesc-S is pretrained on the identical $\sim$18k CodeNet C corpus used by TRACED, with identical data volume, language, backbone, and fine-tuning protocol, so the supervision signal (synthetic descriptions vs.\ execution traces) is the only variable. At matched data the two are on par: clone MAP 0.855 vs.\ 0.858, vulnerability F1 0.557 vs.\ 0.545, title matching 0.624 vs.\ 0.609, tag matching 0.743 vs.\ 0.750, and tag classification 0.220 vs.\ 0.281 (better on two tasks, worse on one, comparable on two), while requiring no execution infrastructure. The headline gains then come from scaling the descriptions (SyncDesc-2S: +6.8 title, +3.1 tag matching, +0.6 tag F1 over TRACED), a scaling path that trace collection does not offer cheaply (Appendix~\ref{app:generation}). For Java, the cleanest pair is UnixCoder vs.\ SyncDesc-1S, initialized from it and differing only by the continued contrastive stage: +9.7 docstring matching, +7.4 CodeChef tag matching, +4.7 description generation.

\subsection{Scale comparison}

We compare our approach with large-scale models on both tag classification and retrieval-based tasks. 
To ensure a fair comparison, we adapt the evaluation protocol to the capabilities of each model family.
For classification, we evaluate closed-source large language models (GPT-family and Claude) in a prompted zero-shot setting, explicitly constraining the output space by providing the list of admissible tags and requiring the model to select from this set. For retrieval-based tasks, we evaluate large-scale models in an embedding-based setting by extracting vector representations and selecting the most similar candidates, mirroring the protocol used for our approach. In this setting, we consider different closed/open-source embedding models as well as DeepSeek-Coder, a generalist open source model. We do not include large-scale model comparisons for vulnerability detection and description generation, as discussed above, since these tasks require fine-grained structural or generative supervision
that is not the primary focus of our contrastive pretraining objective. 

\paragraph{\textbf{Classification:}} \textbf{Constrained multi-class classification over closely related algorithmic tags benefits more from explicitly learned semantic representations than from generative reasoning alone.} Figure~\ref{fig:hero} reports classification performance under identical evaluation settings.
The best-performing model is \textbf{SyncDesc}, which outperforms all prompted large language models,
including Claude Sonnet and GPT-4o, despite being over two orders of magnitude smaller (125M parameters) (F1 0.287 vs 0.270). This comparison confounds adaptation capacity with representation quality: SyncDesc is fine-tuned on task-specific data, whereas the prompted LLMs are zero-shot; the result should be read as a fine-tuned 125M encoder matching zero-shot models two orders of magnitude larger, not as superior representation capability. Claude models consistently outperform GPT-family models, but both remain below the dual-source contrastively pretrained encoder, as well as TRACED execution-aware. Prompted LLMs appear less reliable in separating fine-grained algorithmic categories when forced into a closed-label decision space. The gain is nonetheless not mere output distillation of the description generator: the 125M student exceeds its own teacher on this task (F1 0.287 vs.\ GPT-4o's 0.208), which output imitation alone cannot achieve; swapping the contrastive objective for MLM on identical synthetic data drops 4.2/3.2 points (Table~\ref{tab:ablations}). 

\paragraph{\textbf{Retrieval:}} \textbf{Smaller models with dedicated pre-training can perform similarly or outperform large-scale embedding models, doing so at a fraction of the cost.}
Figure~\ref{fig:hero} reports retrieval performance under matched embedding-based evaluation protocols. Across clone detection, title matching, and tag retrieval, SyncDesc is competitive with or outperforms large-scale embedding models. Against DeepSeek Coder, all pretrained methods achieve substantially higher performance, likely because its autoregressive objective is not designed to produce globally comparable embeddings for fine-grained similarity search. Against Nomic Embed Code and the broader set of embedding LLMs (Qwen3-Embedding, OpenAI text-embedding-3-large, Gemini Embedding-001), SyncDesc achieves higher accuracy on title and tag retrieval while trailing on docstring matching, consistent with these LLM embedders being trained primarily on docstring-style data. Notably, SyncDesc achieves this with a model approximately $56\times$ smaller than Nomic Embed Code, highlighting the scale efficiency of semantically enriched contrastive pretraining. This family is matched: as-is usage is the standard regime for embedding models, and all systems share identical candidate pools and cosine ranking. Table~\ref{tab:zeroshot} (Appendix~\ref{app:detailed_comparison}) further removes the fine-tuning asymmetry within the small-model family by evaluating all 125M encoders frozen: on language-matched tasks SyncDesc retains 88--89\% of its fine-tuned performance and leads the frozen 125M models; cross-language arms collapse and fine-tuning erases these differences.

% \begin{tcolorbox}[
%     colback=white, % Background color
%     colframe=black, % Border color
%     fonttitle=\bfseries,
%     title=Findings RQ2,
%     % arc=4pt, % Rounded corners
%     % left=6pt,
%     % right=6pt,
%     % top=6pt,
%     % bottom=6pt,
%     % boxrule=1pt, % Border thickness
%     ]
% \textbf{Classification:} On closed-label tag classification, the dual-source variant achieves
% the best performance, outperforming all prompted LLMs, including Claude Sonnet and GPT-4o,
% despite being over two orders of magnitude smaller. 
% % Vulnerability detection is excluded from this comparison, as our approach underperforms standard baselines on this task. 

% \textbf{Retrieval:} The proposed method is competitive with or outperforms large-scale embedding models across clone detection, title matching, and tag retrieval.

% \textbf{Objective alignment:} Differences across large models follow their pretraining objectives:
% Nomic Embed Code remains strongest on docstring matching, while contrastive semantic pretraining excels on algorithmic retrieval and classification.
% \end{tcolorbox}

% ======================================================
% SENSITIVITY
% ======================================================
\subsection{Ablation and latent-space analysis}

For all ablation and latent space analyses, we restrict evaluation to tasks where human written descriptions are available but are not part of the evaluation objective. This last point means we had to exclude Docstring Matching (CodeSearchNet), as the task objective is to directly evaluate code--docstring alignment with human docstrings. This would bias the ablation study, as the supervision source is the same as the evaluation target. We therefore conduct ablations on Clone Detection and Tag Classification (Java), where description supervision can be varied without affecting the evaluation signal, and additionally on CodeChef Tag Matching, a code-to-text retrieval task whose target texts (algorithmic tag names) are not the supervision source of any variant; this tests text--code alignment directly on a language-matched target (a C++ target would reintroduce the cross-language confound of Section~\ref{sec:results}).

% ======================================================
% ABLATIONS
% ======================================================

\begin{figure*}[t]
  \centering
  \includegraphics[width=0.75\textwidth]{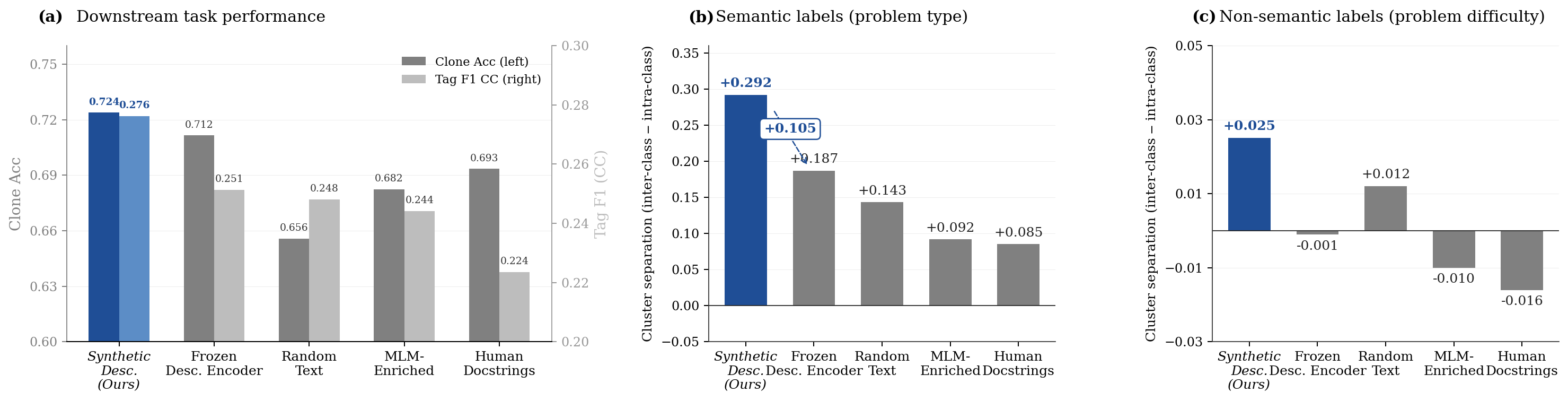}
  \caption{Ablation across alternative supervision sources and training objectives. (a) Downstream-task performance on Clone Accuracy (left axis, dark) and Tag F1 CodeChef (right axis, light); the two metrics use independent y-axis scales to highlight differences within each metric. (b) Cluster separation under semantic algorithmic-category labels. (c) Cluster separation under non-semantic difficulty labels, included as a control. SyncDesc (synthetic descriptions + contrastive objective) achieves the best score across all three panels; alternatives degrade both downstream performance and embedding geometry to varying degrees. Standard deviations and full per-variant numerical results are reported in Appendix~\ref{app:detailed_ablation}.}
  \label{fig:cluster_separation}
\end{figure*}

\subsubsection{\textbf{Ablation Study}} \textbf{Every ablated variant degrades performance, confirming that both the synthetic descriptions and the contrastive objective are necessary.} We ablate each component of SyncDesc by repeating pretraining and fine-tuning with a single element changed (Figure~\ref{fig:cluster_separation} (a)).

As shown in Figure~\ref{fig:cluster_separation}(a), every ablation degrades both Clone Accuracy and Tag F1 relative to SyncDesc (0.724 and 0.276). Semantic content (random text) and substituting MLM for contrastive learning, are the primary drivers of gains. The same ordering extends to retrieval: on CodeChef Tag Matching the full model leads every ablated arm by at least 2.5 points, with human docstrings again last, below random text (Table~\ref{tab:ablation_ccmatch}). Full results are reported in Appendix~\ref{app:detailed_ablation}.

\paragraph{\textbf{Supervision-source controls:}} \textbf{Semantic extraction from the code, not description writing quality or the specific generator, drives the effect.} On an identical 50k subset of the 1S corpus where only the supervision text varies (Table~\ref{tab:supervision_source}, Appendix~\ref{app:detailed_ablation}), GPT-4o paraphrases of the human docstrings produced \emph{without access to the code} recover none of the human-to-synthetic gap (clone accuracy 0.728 vs.\ 0.752 for code-conditioned synthetic; tag F1 0.230 vs.\ 0.260), and replacing GPT-4o with the open-weights Qwen2.5-Coder-7B as generator matches within seed noise, so the effect is not GPT-4o-specific and generation cost can drop to essentially zero.

Beyond the ablation, we further analyze the properties of the synthetic descriptions themselves. Synthetic descriptions are roughly 3$\times$ longer than human docstrings, exhibit substantially higher lexical diversity, and reuse code identifiers far less often (see Appendix~\ref{app:descprops}, Table~\ref{tab:lexical_analysis}). A sensitivity analysis across description lengths (25, 50, 100 tokens) shows no monotonic relationship between length and downstream performance (see Appendix~\ref{app:descprops}, Table~\ref{tab:token_sensitivity}), indicating that abstraction quality matters more than raw length. We return to these properties when interpreting the latent-space analysis below.

% ======================================================
% REPRESENTATION ANALYSIS
% ======================================================
\paragraph{\textbf{Latent space analysis:}} \textbf{Synthetic docstring with contrastive learning improve embedding in small transformer.} To assess how human and synthetic docstrings are represented in the latent space, we measure the average cluster separation (inter-class minus intra-class distance) of their embeddings, using algorithmic category tags (e.g., sorting) and difficulty labels from the CodeChef (Java) test set, and compare against embeddings trained with a frozen description encoder and random docstrings, as in our ablation study. Results in Figure~\ref{fig:cluster_separation} (b, c) confirm that our approach (synthetic docstrings + contrastive pre-training) achieves a higher separation than the frozen encoder on \textit{Problem type} tags (panel~(b), 0.292 vs 0.187), and a higher separation than random text on \textit{Problem Difficulty} (panel~(c), 0.025 vs 0.012). Human docstrings generally underperform even on \textit{Problem Type}, where the semantics of the tasks should be more evident. In contrast, our approach improves separation, highlighting the stronger semantic embedding, even on \textit{Problem Difficulty}, where semantic alignment might be harder to achieve. The weaker clustering observed for human-written docstrings can be explained by their lexical and structural properties. We show in Appendix~\ref{app:descprops} (Table~\ref{tab:lexical_analysis}) that human docstrings exhibit substantially higher identifier overlap and lower lexical diversity than synthetic descriptions, indicating a tendency to reuse surface-level code tokens rather than abstract functional intent. Moreover, their large variance in length and content leads to inconsistent levels of semantic abstraction across samples. In a contrastive embedding space, such properties reduce the ability to form stable semantic anchors, resulting in blurred cluster boundaries and lower inter-class separation~\cite{chen2020simple, wang2020understanding}. In contrast, synthetically generated descriptions enforce consistent abstraction and vocabulary diversity, which promotes tighter grouping of semantically related programs, as reflected in the improved separation scores in Figure~\ref{fig:cluster_separation}(b,c). A complementary UMAP visualization is provided in Appendix~\ref{app:umap}.

% \Florian{Feel free to add, I just did something simple}
% \begin{tcolorbox}[
%     colback=white, % Background color
%     colframe=black, % Border color
%     fonttitle=\bfseries,
%     title=Findings RQ3,
%     % arc=4pt, % Rounded corners
%     % left=6pt,
%     % right=6pt,
%     % top=6pt,
%     % bottom=6pt,
%     % boxrule=1pt, % Border thickness
%     ]
%     \textbf{Abaltion Study:} Contrastive pre-training and synthetic descriptions both contribute to the performance of the approach, validating their usefulness.
    
%     \textbf{Synthetic vs Human docstring:} Synthetic docstrings differs from human docstrings: the content of the docstring is what matters most, with synthetic docstrings having better semantic properties.
    
%     \textbf{Latent space embeddings:} Synthetic docstrings leads to more meaningful embeddings, due to better clustering of semantically similar tasks compared to alternative. 
% \end{tcolorbox}

% \input{sections/discussion}

\section{Related Work}

\subsection{Models for code embeddings and downstream tasks}

Early pretrained code encoders model source code as token sequences using denoising-style objectives. Models such as CodeBERT~\cite{feng2020codebert} and GraphCodeBERT~\cite{guo2020graphcodebert} combine masked language modeling with auxiliary signals, including replaced token detection and data-flow modeling, while encoder–decoder architectures like UnixCoder~\cite{guo2022unixcoder} and CodeT5~\cite{wang2021codet5} extend this paradigm to support both understanding and generation via multi-objective pretraining. Despite strong performance, these methods achieve only indirect code–text semantic alignment, as supervision relies on reconstruction or generation losses and the quality of naturally occurring comments or documentation. Execution-aware approaches such as TRACED~\cite{ding2024traced} incorporate dynamic execution information (e.g., traces, coverage, runtime values), but require executable code, runtime infrastructure, and dataset-specific tooling, which limits scalability. Large autoregressive code LLMs attain strong generative capabilities through massive-scale next-token prediction, as exemplified by GPT-4o-mini~\cite{openai_gpt4o_mini_2024}, while embedding-oriented models like Nomic Embed Code~\cite{nussbaum2025nomic} train dual encoders contrastively on large code–docstring corpora for retrieval. Although these approaches benefit from scale and data volume, their objectives remain primarily generative or broadly textual and may underperform on specialized tasks. \textit{In contrast}, we propose a synthetic docstring–based contrastive learning approach that improves upon classical semantic alignment as in UnixCoder and scales more effectively than both execution-aware methods and LLMs while achieving competitive performance.

\subsection{Contrastive Code--Text Alignment}

Contrastive learning was first popularized for image--text alignment~\cite{radford2021clip} and has since expanded considerably~\cite{hu2024comprehensive}. In software engineering, it has also emerged as a powerful mechanism for improving code--text alignment, often integrated as a secondary objective in broader pretraining frameworks. For example, UnixCoder and CodeT5+ \cite{wang2023codet5+} incorporate contrastive objectives alongside masked or generative losses. However, these approaches do not treat contrastive alignment as the primary training signal. Dual-encoder formulations are common in retrieval settings~\cite{husain2019codesearchnet}, but their performance is constrained by the variability and semantic fidelity of developer-written comments or metadata~\cite{sun2022importance}. Techniques like hard-negative mining~\cite{robinson2020contrastive} partially address these issues, but alignment quality remains limited by the supervision source. When contrastive alignment is treated as a secondary objective, representation geometry remains largely shaped by reconstruction or generation losses, with contrastive signals acting as a weak regularizer. In contrast, making contrastive alignment the primary training signal directly optimizes the embedding space for semantic similarity and discrimination, which is more aligned with retrieval and matching-based downstream use~\cite{wang2020understanding}. Datasets such as FunCom~\cite{bansal2023funcom}, DeepCom~\cite{hu2020deepcom}, and CodeSearchNet~\cite{husain2019codesearchnet} provide large-scale natural-language supervision and underpin prior work in code summarization
and retrieval. Their annotations are typically short fragments derived from docstrings or
headers and exhibit substantial variability in style, completeness, and
abstraction.

A related line of work explores contrastive learning for code through different design axes. ContraCode~\cite{jain2021contrastive}, Corder~\cite{bui2021corder}, SynCoBERT~\cite{wang2021syncobert}, and Concord~\cite{ding2023concord} study code-specific augmentations such as variable renaming, dead-code insertion, or other semantics-preserving transformations. SimCSE~\cite{gao2021simcse} uses dropout-based augmentation for general sentence embeddings. CLEAR~\cite{wei2022clear} and CoCoSoDa~\cite{shi2023cocosoda} propose retrieval-specific architectures for code search. These works vary the contrastive mechanism (augmentation strategy, dropout, hard negatives) or the architecture. Our contribution is orthogonal: we hold both constant and vary the supervision source, replacing augmentations and human docstrings with synthetic semantic descriptions. The two design dimensions are complementary, and combining synthetic semantic supervision with the augmentation or architectural advances above is a future work direction.

This work treats contrastive alignment as the central objective and pairs code with \emph{synthetic semantic descriptions} designed to provide consistent, instance-level grounding, without relying on potentially missing or low-quality docstrings.

\label{sec:related_works}

\section{Conclusion}

This study compared code-embedding methods' performance and studied whether a simple approach leveraging synthetically generated semantic descriptions can effectively supervise the learning of general-purpose code representations. The approach, SyncDesc, is a lightweight knowledge distillation: a large teacher's semantic extraction is transferred into a compact encoder and the encoder then serves embedding workloads, effectively keeping the advantage of zero-shot generalization. SyncDesc is evaluated on eight downstream coding tasks (spanning retrieval, classification, and generation) against several transformer baselines across C, C++, and Java, demonstrating better performance on five of the eight tasks. Performance on structurally oriented tasks, such as vulnerability detection, remains comparable to existing baselines, suggesting that this approach complements rather than replaces dynamic supervision. The results indicate that semantically consistent contrastive supervision can produce compact, competitive code representations, matching more complex baselines at matched pretraining data and surpassing them when descriptions are scaled, in semantic tasks while maintaining effectiveness across diverse evaluations. 
Comparisons with larger models highlight that small scale encoders trained with synthetic semantic supervision can achieve competitive performance, after task-specific fine-tuning that the zero-shot LLMs lack, emphasizing the importance of supervision quality and training objectives, particularly under efficiency and deployability constraints.

Future work could extend synthetic supervision to more programming languages and multi-file or project-level code, analyze how description properties (e.g., abstraction level) interact with downstream tasks, and integrate structured signals (e.g., type information) to enhance representations for tasks requiring deeper semantic or structural reasoning. These directions may further strengthen the potential of synthetic semantic supervision in learning robust, transferable code representations. 

\label{sec:conclusion}

\clearpage

% This is specific to EMNLP. Does not count towards 8 pages limit
\section*{Limitations}

Our supervision signal is generated by GPT-4o and may reflect generation artifacts (omitted semantics or stylistic biases) rather than purely generalizable semantic structure. We mitigate this with a fixed prompting template applied uniformly across datasets and languages, but the description quality is not independently verifiable. Exact regeneration also depends on GPT-4o, which may evolve over time.

We evaluate on method-level code across three languages (C, C++, Java), drawn from competitive-programming corpora (CodeNet, CodeForces, CodeChef) and GitHub-mined repository code (FunCom, DeepCom, CONCODE, CodeSearchNet, BigCloneBench, Devign). All training and evaluation nonetheless operate on isolated methods: industrial, multi-file, and project-level distributions (functions whose semantics depend on surrounding modules, frameworks, or build context, and repository-level retrieval) remain untested, as do other languages.

Our 125M-scale baselines are fine-tuned per task, while LLM-scale embedding and prompted models are evaluated as-is, reflecting typical practice but not matched adaptation effort. A frozen-encoder evaluation of all 125M models without any fine-tuning (Table~\ref{tab:zeroshot}) partially removes this asymmetry within the small-model family. We strived to make the comparison across baselines as fair as possible, yet difference in pre-training dataset or objectives might have affected performance on the studied tasks. The ContraCode language confound and the confound-free matched-data comparisons are discussed directly in Section~\ref{sec:results}.

\section*{Acknowledgments}

This work is supported by the Luxembourg National Research Fund (FNR) through the CORE project C23/IS/18182513/MiCE.

\clearpage

\bibliography{custom}

\clearpage

\appendix

\section{Full results with standard deviations}
\label{app:full_results}

This appendix reports the complete numerical results summarized in Figure~\ref{fig:hero}, including all metrics (e.g., similarity separation) and per-seed standard deviations omitted from the figure for readability. Table~\ref{tab:significance} additionally reports paired instance-level bootstrap tests for every headline comparison against the strongest 125M baseline: improvements are statistically significant on five of the eight tasks, better on average but not significantly on two (description generation and CodeForces tag classification), and at parity-to-below on vulnerability detection, where SyncDesc-S trails UnixCoder (n.s.) while exceeding TRACED at matched pretraining data (0.557 vs.\ 0.545).

\begin{table*}[t]
  \centering
  % \scriptsize
  \setlength{\tabcolsep}{4pt}
  \caption{Results for mid-scale (125M) models across all downstream tasks. Metrics are averaged over three seeds with standard deviation. $-S$ = single source, $-2S$ = dual source. For $-S$, for each downstream task, we used the pre-train dataset matching the programming language of the downstream task. We give in \textbf{bold} (resp. \underline{underlined}) the best (resp. second best) approach per task. $^\dagger$ marks SyncDesc results whose improvement over the strongest 125M baseline is statistically significant under paired instance-level bootstrap tests (10k resamples, 95\% CIs on the metric difference; Table~\ref{tab:significance}).}
  \label{tab:rq1_main}
  \resizebox{\textwidth}{!}{%
  \begin{tabular}{l|c|cc|c|c|cc|cc|cc|c}
    \toprule
    \textbf{Model} &
    \textbf{Clone} &
    \multicolumn{2}{c|}{\textbf{Vuln.}} &
    \textbf{Title Match (CF)} &
    \textbf{Tag Classif. (CF)} &
    \multicolumn{2}{c|}{\textbf{Tag Match (CF)}} &
    \multicolumn{2}{c|}{\textbf{Tag Match (CC)}} &
    \multicolumn{2}{c|}{\textbf{Docstr. match}} &
    \multicolumn{1}{c}{\textbf{Docstr. gen}} \\
    &
    MAP $\uparrow$ &
    Acc $\uparrow$ & F1 $\uparrow$ &
    Acc $\uparrow$ &
    F1 $\uparrow$ &
    Acc $\uparrow$ & Sep $\uparrow$ &
    Acc $\uparrow$ & Sep $\uparrow$ &
    Acc $\uparrow$ & Sep $\uparrow$ &
    Sim $\uparrow$ \\
    \midrule
    % CB baseline  & 0.803 $\pm$ 0.030 & 0.637 $\pm$ 0.005 & 0.541 $\pm$ 0.005 & 0.576 $\pm$ 0.016 & 0.173 $\pm$ 0.030 & 0.641 $\pm$ 0.059 & 0.346 $\pm$ 0.319 & 0.682 $\pm$ 0.051 & 1.693 $\pm$ 0.319 & 0.506 $\pm$ 0.049 & 1.757 $\pm$ 0.308 & 0.603 $\pm$ 0.005 \\
    UnixCoder & 0.857 $\pm$ 0.012 & \textbf{0.651 $\pm$ 0.008} & \textbf{0.570 $\pm$ 0.007} & 0.607 $\pm$ 0.048 & 0.232 $\pm$ 0.025 & 0.743 $\pm$ 0.005 & 1.397 $\pm$ 0.096 & 0.636 $\pm$ 0.030 & 1.632 $\pm$ 0.096 & 0.548 $\pm$ 0.001 & 1.866 $\pm$ 0.088 & 0.625 $\pm$ 0.055 \\
    ContraCode  & 0.803 $\pm$ 0.006 & 0.632 $\pm$ 0.007 & \underline{0.568} $\pm$ 0.019 & 0.524 $\pm$ 0.005 & 0.227 $\pm$ 0.006 & 0.603 $\pm$ 0.008 & 1.571 $\pm$ 0.111 & 0.655 $\pm$ 0.008 & \textbf{2.369 $\pm$ 0.083} & 0.595 $\pm$ 0.031 & 1.378 $\pm$ 0.372 & 0.658 $\pm$ 0.063 \\
    TRACED             & \underline{0.858 $\pm$ 0.024} & 0.641 $\pm$ 0.002 & 0.545 $\pm$ 0.002 & 0.609 $\pm$ 0.044 & \underline{0.281 $\pm$ 0.031} & \underline{0.750 $\pm$ 0.012} & 1.395 $\pm$ 0.113 & \underline{0.663 $\pm$ 0.004} & \underline{1.995} $\pm$ 0.047 & 0.568 $\pm$ 0.008 & 1.636 $\pm$ 0.113 & 0.648 $\pm$ 0.004 \\
    \midrule
    % Our (CB, 1S) & 0.819 $\pm$ 0.015 & 0.635 $\pm$ 0.001 & \textbf{0.585 $\pm$ 0.016} & 0.561 $\pm$ 0.064 & 0.211 $\pm$ 0.001 & 0.658 $\pm$ 0.012 & \textbf{2.602 $\pm$ 0.216} & \underline{0.696 $\pm$ 0.036} & \underline{1.873 $\pm$ 0.216} & 0.633 $\pm$ 0.009 & \underline{2.366 $\pm$ 0.016} & 0.652 $\pm$ 0.018 \\
    SyncDesc-S &
                      0.855 $\pm$ 0.009 & \underline{0.646 $\pm$ 0.002} & 0.557 $\pm$ 0.007 & \underline{0.624 $\pm$ 0.062} & 0.220 $\pm$ 0.012 & 0.743 $\pm$ 0.003 & \underline{1.690} $\pm$ 0.201 & \textbf{0.710$^\dagger$ $\pm$ 0.049} & 1.842 $\pm$ 0.201 & \textbf{0.645$^\dagger$ $\pm$ 0.011} & \underline{2.283} $\pm$ 0.051 & \textbf{0.672 $\pm$ 0.001} \\
    SyncDesc-2S &
                      \textbf{0.870$^\dagger$ $\pm$ 0.014} & 0.6354 $\pm$ 0.001 & 0.5225 $\pm$ 0.014 & \textbf{0.677$^\dagger$ $\pm$ 0.005} & \textbf{0.287 $\pm$ 0.029} & \textbf{0.781$^\dagger$ $\pm$ 0.004} & \textbf{1.834 $\pm$ 0.158} & 0.650 $\pm$ 0.045 & 1.547 $\pm$ 0.158 & \underline{0.641 $\pm$ 0.007} & \textbf{2.525 $\pm$ 0.056} & \underline{{0.670 $\pm$ 0.009}}\\
    \bottomrule
  \end{tabular}  
  }
  
\end{table*}

\begin{table*}[t]
  \centering
  \small
  \setlength{\tabcolsep}{5pt}
  \caption{Paired instance-level bootstrap significance of the headline comparisons in Table~\ref{tab:rq1_main}: SyncDesc (per-task headline variant, as labeled) against the strongest 125M baseline on every task, over saved per-item predictions (10k resamples; 95\% CI on the metric difference).}
  \label{tab:significance}
  \begin{tabular}{l l l r c l}
    \toprule
    \textbf{Task} & \textbf{SyncDesc (variant)} & \textbf{Strongest 125M baseline} & \textbf{$\Delta$ (pts)} & \textbf{95\% CI} & \textbf{Verdict} \\
    \midrule
    CF Title Match        & 0.677 (2S) & 0.609 (TRACED)     & $+6.8$ & $[+5.2, +8.4]$ & significant \\
    Docstring Match       & 0.645 (S)  & 0.595 (ContraCode) & $+5.0$ & $[+4.4, +5.6]$ & significant \\
    CC Tag Match          & 0.710 (S)  & 0.663 (TRACED)     & $+4.7$ & $[+2.9, +6.5]$ & significant \\
    CF Tag Match          & 0.781 (2S) & 0.750 (TRACED)     & $+3.1$ & $[+1.6, +4.6]$ & significant \\
    Desc.\ Generation     & 0.672 (S)  & 0.658 (ContraCode) & $+1.4$ & $[-0.4, +3.2]$ & not significant \\
    Clone MAP (POJ-104)   & 0.870 (2S) & 0.858 (TRACED)     & $+1.2$ & $[+0.1, +2.5]$ & significant \\
    CF Tag-Clf F1         & 0.287 (2S) & 0.281 (TRACED)     & $+0.6$ & $[-1.9, +3.1]$ & not significant \\
    Vulnerability F1      & 0.557 (S)  & 0.570 (UnixCoder)  & $-1.3$ & $[-3.1, +0.5]$ & not significant (below) \\
    \bottomrule
  \end{tabular}
\end{table*}

\begin{table*}[t]
  \centering
  \tiny
  \caption{Code function classification (CodeForces Top-10 Tags). Weighted F1 averaged across three seeds (std. in parentheses). We give in \textbf{bold} (resp. \underline{underlined}) the best (resp. second best) approach per task. Prompted LLMs are evaluated zero-shot with a constrained label set; SyncDesc variants are fine-tuned on task data, so the comparison reflects adaptation regime as well as representation quality (Section~\ref{sec:results}).}
  \label{tab:rq3_classification}
  \begin{tabular}{l r}
    \toprule
    \textbf{Model} & \textbf{F1} \\
    \midrule
    GPT-4o-mini     & 0.151 ($\pm$ 0.018) \\
    GPT-3.5-Turbo  & 0.199 ($\pm$ 0.021) \\
    GPT-4o         & 0.208 ($\pm$ 0.017) \\
    Claude Haiku   & 0.234 ($\pm$ 0.024) \\
    Claude Sonnet  & \underline{0.270 ($\pm$ 0.019)} \\
    \midrule
    SyncDesc-S & 0.220 ($\pm$ 0.012) \\
    SyncDesc-2S   & \textbf{0.287 ($\pm$ 0.029)} \\
    \bottomrule
  \end{tabular}
\end{table*}

\begin{table*}[t]
  \centering
  \tiny
  \caption{Scale comparison under matched retrieval evaluation protocols. Metrics are averaged over three seeds (std. in parentheses). Superscripts on the SyncDesc row indicate the reporting variant per task ($S$ = single-source, $2S$ = dual-source), matching Table~\ref{tab:rq1_main}. All 125M models are fine-tuned per task; the LLM-scale embedders are used as-is, their standard deployment regime, under identical candidate pools and cosine ranking.}
  \label{tab:scale_comparison}
  \resizebox{\textwidth}{!}{\begin{tabular}{l c c c c c}
    \toprule
    \textbf{Model} &
    \textbf{Clone MAP@499} &
    \textbf{Title Match Acc. (CF)} &
    \textbf{Tag Match Acc. (CF)} &
    \textbf{Tag Match Acc. (CC)} &
    \textbf{Docst. match Acc.} \\
    \midrule
    SyncDesc &
    0.870$^{2S}$ ($\pm$ 0.014) &
    \textbf{0.677$^{2S}$ ($\pm$ 0.005)} &
    \textbf{0.781$^{2S}$ ($\pm$ 0.004)} &
    \textbf{0.710$^{S}$ ($\pm$ 0.049)} &
    0.645$^{S}$ ($\pm$ 0.011) \\
    Nomic Embed Code (7B)    &
    0.908 ($\pm$ 0.008) &
    0.629 ($\pm$ 0.012) &
    0.570 ($\pm$ 0.015) &
    \underline{0.660 ($\pm$ 0.021)} &
    0.985 ($\pm$ 0.003) \\
    DeepSeek Coder (6.7B)    &
    0.018 ($\pm$ 0.009) &
    0.486 ($\pm$ 0.018) &
    0.475 ($\pm$ 0.023) &
    0.443 ($\pm$ 0.027) &
    0.700 ($\pm$ 0.011) \\
    Qwen3-Embedding (0.6B)    &
    \underline{0.955 ($\pm$ 0.001)} &
    0.588 ($\pm$ 0.001) &
    \underline{0.600 ($\pm$ 0.009)} &
    0.557 ($\pm$ 0.002) &
    \textbf{0.992 ($\pm$ 0.001)} \\
    OpenAI text-embedding-3-large    &
    0.713 ($\pm$ 0.002) &
    \underline{0.664 ($\pm$ 0.029)} &
    0.539 ($\pm$ 0.048) &
    0.401 ($\pm$ 0.007) &
    \underline{0.992 ($\pm$ 0.001)} \\
    Gemini embedding-001    &
    \textbf{0.994 ($\pm$ 0.000)} &
    0.612 ($\pm$ 0.004) &
    0.595 ($\pm$ 0.010) &
    0.502 ($\pm$ 0.003) &
    0.982 ($\pm$ 0.001) \\
    \bottomrule
  \end{tabular}}
\end{table*}

\section{Detailed downstream-task comparison}
\label{app:detailed_comparison}

This appendix expands the condensed comparison in Section~\ref{sec:results} with the full per-task breakdown and the single- versus dual-source analysis.

\paragraph{\textbf{Comparison to standard baselines and execution-aware pretraining.}} Table~\ref{tab:rq1_main} reports results across all downstream tasks. For single-source, for each downstream task, we used the pre-train dataset matching the programming language of the downstream task. Across retrieval, matching, and classification tasks, models pretrained with synthetic semantic descriptions consistently match or improve upon standard transformer baselines and the execution-aware model, with the improvements statistically significant on five of eight tasks (Table~\ref{tab:significance}). The strongest improvements are observed on semantically oriented retrieval tasks, including clone retrieval (MAP 0.870 for our approach vs 0.858 strongest baseline), title matching (Accuracy 0.677 vs 0.609), tag matching (Accuracy 0.781 vs 0.750 on CodeForces and 0.710 vs 0.663 on CodeChef), and docstring matching (Accuracy 0.645 vs 0.595), where our method attains the highest score across all metrics; all four gains are significant. Improvements are particularly pronounced in similarity separation rather than ranking accuracy, indicating better discrimination between matched and non-matched items. On tag classification, synthetic semantic pretraining is comparable to the strongest baseline on CodeForces (F1 0.287 vs 0.281; not statistically significant). On vulnerability detection, performance differences across models are small and remain within a narrow range, which is consistent with the task's reliance on fine-grained control-flow and implementation-level patterns that are not directly emphasized by semantic-level supervision (F1 0.557 vs 0.570; n.s., and above TRACED at matched pretraining data, 0.557 vs 0.545). For description generation, models pretrained with synthetic semantic supervision achieve the highest semantic similarity scores (Similarity 0.672 vs 0.658 for ContraCode, the strongest baseline; $+1.4$ points, not statistically significant).

\paragraph{\textbf{Single-source versus dual-source pretraining.}}
Table~\ref{tab:rq1_main} shows that the effect of dual-source pretraining depends strongly on the relative richness of the single-source corpus. For C/C++ tasks (clone retrieval on POJ-104 and CodeForces title, tag matching, and tag classification), single-source pretraining relies exclusively on CodeNet, which provides samples from a single benchmark-style corpus. In this setting, adding Java data (SyncDesc-2S) yields consistent gains over C-only pretraining, e.g., +1.5 MAP on clone retrieval (0.870 vs.\ 0.855) and +5.7 accuracy points on CodeForces title matching (0.677 vs.\ 0.624). In contrast, Java-centric tasks are pretrained on a substantially larger and more heterogeneous Java corpus aggregating FunCom, DeepCom, and CONCODE. For these tasks, single-source Java pretraining already provides strong coverage, and adding C data yields no systematic benefit, with performance remaining comparable or slightly lower (e.g., CodeChef tag matching and docstring matching). These results indicate that source mixing primarily compensates for limited single-source coverage, rather than providing uniform gains from increased data volume alone.

\paragraph{\textbf{Zero-shot frozen-encoder evaluation.}}\label{app:zeroshot}
To assess transfer without any task-specific fine-tuning, Table~\ref{tab:zeroshot} evaluates all 125M models with frozen, mean-pooled embeddings under the identical protocol used for the embedding-model rows of Table~\ref{tab:scale_comparison} (hence these numbers are not comparable to the fine-tuned results in Table~\ref{tab:rq1_main}). On language-matched tasks, the pretrained encoder transfers directly: SyncDesc-1S-C reaches 0.752 zero-shot clone MAP ($+24.6$ points over frozen UnixCoder; 88\% of its own fine-tuned 0.855), and SyncDesc-1S-Java is best among frozen 125M models on CodeChef tag matching (0.632, about 89\% of its fine-tuned 0.710, marginally above TRACED's 0.631). We restrict the transfer claim to language-matched settings: cross-language arms collapse on clone retrieval (1S-Java: 0.258), i.e., language match dominates zero-shot embedding geometry, and fine-tuning erases these differences (Table~\ref{tab:rq1_main}). The CodeForces columns are flat across all frozen 125M models, where the large embedding LLMs retain their lead (Table~\ref{tab:scale_comparison}). The docstring column is led by vanilla UnixCoder; under this frozen protocol the task is lexically easy (cf.\ the 0.98+ rows for embedding LLMs in Table~\ref{tab:scale_comparison}), and we draw no transfer claim from it.

\begin{table*}[t]
  \centering
  \small
  \setlength{\tabcolsep}{5pt}
  \caption{Zero-shot frozen-encoder retrieval. Mean-pooled embeddings, no fine-tuning; identical protocol to the embedding-model rows of Table~\ref{tab:scale_comparison}. Clone and CodeChef tag matching use the full test sets (deterministic); the docstring and CodeForces columns use 10k samples over 3 seeds.}
  \label{tab:zeroshot}
  \begin{tabular}{l c c c c c}
    \toprule
    \textbf{Model (125M, frozen)} &
    \textbf{Clone MAP} &
    \textbf{Docstr.\ Match} &
    \textbf{CF Title} &
    \textbf{CF Tags} &
    \textbf{CC Tags} \\
    \midrule
    UnixCoder        & 0.506 & \textbf{0.945} $\pm$0.003 & 0.524 $\pm$0.011 & 0.531 $\pm$0.009 & 0.613 \\
    TRACED           & 0.346 & 0.877 $\pm$0.004 & 0.511 $\pm$0.003 & 0.514 $\pm$0.007 & \underline{0.631} \\
    ContraCode       & 0.345 & 0.907 $\pm$0.002 & \underline{0.527} $\pm$0.007 & 0.495 $\pm$0.003 & 0.498 \\
    SyncDesc-1S-Java & 0.258 & \underline{0.915} $\pm$0.001 & 0.514 $\pm$0.008 & \underline{0.515} $\pm$0.007 & \textbf{0.632} \\
    SyncDesc-1S-C    & \textbf{0.752} & 0.918 $\pm$0.000 & \textbf{0.534} $\pm$0.008 & \textbf{0.534} $\pm$0.008 & 0.469 \\
    SyncDesc-2S      & \underline{0.561} & 0.886 $\pm$0.001 & 0.531 $\pm$0.004 & 0.509 $\pm$0.007 & 0.446 \\
    \bottomrule
  \end{tabular}
\end{table*}

\section{Detailed ablation study}
\label{app:detailed_ablation}

This appendix provides the per-variant ablation analysis summarized visually in Figure~\ref{fig:cluster_separation}. For our ablation study, we repeat the pre-train and fine-tuning steps, except that we operate several ablations of our approach to evaluate the contribution towards the performance of each part of our approach. Full numerical results with standard deviations are given in Table~\ref{tab:ablations}.

\begin{table*}[t]
  \centering
  \tiny
  \caption{Ablation results across alternative supervision sources and training objectives.}
  \label{tab:ablations}
  \begin{tabular}{l c c}
    \toprule
    \textbf{Variant} &
    \textbf{Clone Acc} &
    \textbf{Tag F1 (CC)} \\ %&
    % \textbf{Docstring Rank Acc} \\
    \midrule
    SyncDesc-S & \textbf{0.7239 $\pm$ 0.016} & \textbf{0.2762 $\pm$ 0.009} \\ %& 0.6451 $\pm$ 0.013 \\ 
    Human Docstrings                     & 0.6934 $\pm$ 0.013 & 0.2235 $\pm$ 0.098 \\ %& \textbf{0.6739 $\pm$ 0.005} \\
    Frozen Description Encoder           & 0.7116 $\pm$ 0.014 & 0.2513 $\pm$ 0.052 \\ %& 0.6636 $\pm$ 0.000 \\
    MLM-Enriched                         & 0.6824 $\pm$ 0.014 & 0.2441 $\pm$ 0.051 \\ %& 0.6024 $\pm$ 0.000 \\
    Random Text                          & 0.6558 $\pm$ 0.011 & 0.2481 $\pm$ 0.026 \\ %& 0.3552 $\pm$ 0.023 \\
    \bottomrule
  \end{tabular}
\end{table*}

\paragraph{\textbf{Human-Docstring}} We first repeated the whole process using available human docstrings instead of our generated synthetic docstrings, with the rest of the process unchanged. The goal was to motivate the necessity of synthetic docstrings. This translated into a drop in performance of, on average, approximately 3 percentage points for Accuracy/Clone detection and approximately 5 percentage points for F1/Tag classification. As such, on top of being scarcely available, native human docstrings generally degrade performance compared to our synthetic docstrings.

\paragraph{\textbf{Random-Text}} Secondly, we repeated the process by replacing synthetic descriptions with random words of the same length. The goal was to show that performance was not just a byproduct of the training methodology. Swapping in random text degraded performance the most on average: performance dropped by approximately 7 percentage points on Clone detection and approximately 3 percentage points on Tag classification. As such, while fine-tuning certainly helps performance despite random text, the performance of our approach is not driven solely by the training process but also by the added synthetic descriptions.

\paragraph{\textbf{Frozen Description Encoder}} Thirdly, we repeated the process by freezing the whole description encoder (DeBERTa base + projection layer) when pre-training our approach. The goal was to assess whether the approach did not just benefit from the code encoder learning, and so that the descriptions pairing helped. This translated into a drop of approximately 1 percentage point for Clone detection and approximately 2 percentage points for Tag classification. In practice, this means that keeping the description encoder frozen appears to have limited impact on tasks, meaning the embedding alignment is compensated mainly by the code encoder. Nonetheless, it still decreases performance on average, at a negligible cost since updating the whole description encoder is cheap given its size.

\paragraph{\textbf{MLM-Enriched}} Finally, we repeated the process by training our approach with masked language modelling (similarly to classical CodeBERT) instead of contrastive learning. The goal was to show that the contrastive objective benefited the approach over the simple masked language modelling task. This resulted in a drop of 4.2 percentage points for Clone detection and 3.2 percentage points for Tag classification. As such, contrastive learning, which uses negative anchoring to train the embedding, seems to improve performance.

\paragraph{\textbf{Ablation on a direct text--code alignment target.}}
The two ablation tasks above are classification tasks; to test the central alignment hypothesis directly, Table~\ref{tab:ablation_ccmatch} repeats the full five-variant ablation on CodeChef Tag Matching, a code-to-text retrieval task whose target texts (algorithmic tag names) are not the supervision source of any variant. We select this task rather than CodeForces title matching because the ablation variants are pretrained on the Java corpus and CodeChef is Java: a C++ target would entangle the supervision-source effect with a cross-language confound. The full model exceeds every ablated arm by at least 2.5 points on a target that no variant was supervised on. Human docstrings again rank last, below even random text, extending the ordering of Table~\ref{tab:ablations} from classification to retrieval; the human-docstring arm also sits close to the plain UnixCoder reference, i.e., human docstrings add little over no continued pretraining here. Because the full model's seed-level standard deviation on this task is large ($\pm 0.049$), we pair this comparison with the instance-level bootstrap protocol of Table~\ref{tab:significance} (1,890 test items), alongside the similarity-separation geometry already reported in Table~\ref{tab:rq1_main} (1.842 vs.\ 1.632 for UnixCoder). We also note that the main results already contain non-docstring text--code alignment evidence: title matching ($+6.8$ points over the best baseline) and tag matching ($+3.1$/$+4.7$ points), where the candidate pools are natural-language titles and tag names.

\begin{table}[t]
  \centering
  \caption{Five-variant ablation on CodeChef Tag Matching (accuracy, 3 seeds, identical fine-tuning protocol). UnixCoder is shown as a reference, not an ablation arm.}
  \label{tab:ablation_ccmatch}
  \resizebox{0.85\columnwidth}{!}{%
  \begin{tabular}{l c}
    \toprule
    \textbf{Variant} & \textbf{CC Tag Match Acc} \\
    \midrule
    SyncDesc-S (synthetic + contrastive) & \textbf{0.710 $\pm$ 0.049} \\
    Frozen Description Encoder           & 0.685 $\pm$ 0.030 \\
    Random Text                          & 0.672 $\pm$ 0.025 \\
    MLM-Enriched                         & 0.668 $\pm$ 0.030 \\
    Human Docstrings                     & 0.655 $\pm$ 0.035 \\
    \midrule
    UnixCoder (reference)                & 0.636 $\pm$ 0.030 \\
    \bottomrule
  \end{tabular}}
\end{table}

\paragraph{\textbf{Controlled supervision-source suite (paraphrase control and open-weights generator).}}
To isolate description \emph{writing quality} from \emph{code-derived semantic content}, and to test sensitivity to the generating model, we run four supervision arms on an identical 50k subset of the 1S corpus, differing only in the supervision text, with the identical pretraining recipe and fine-tune/evaluation protocol as Table~\ref{tab:ablations} (3 seeds). Because this suite uses a 50k subset, its absolute numbers are not comparable to the full-corpus results in Table~\ref{tab:ablations}. Arm (2) has GPT-4o rewrite the human docstrings \emph{without access to the code}, normalizing style, length, and lexical diversity under a no-added-information constraint; arm (4) replicates description generation with the open-weights Qwen2.5-Coder-7B using the paper's prompt and decoding parameters. Results are in Table~\ref{tab:supervision_source}. Rewritten docstrings do not close the gap: clone accuracy 0.728 vs.\ 0.752 for code-conditioned synthetic (raw human: 0.739), tag F1 0.230 vs.\ 0.260 (raw human: 0.229); none of the human-to-synthetic gap is recovered by surface quality alone. Both code-conditioned arms sit above both docstring-derived arms on both tasks, with markedly tighter seed variance: what drives the effect is semantic extraction from the code itself, not surface quality. The Qwen2.5-Coder-7B arm matches GPT-4o within seed noise on both tasks (differences of 0.003 and 0.004), so the effect is not GPT-4o-specific and the one-time generation cost can drop to essentially zero on a local GPU with open weights.

\begin{table}[t]
  \centering
  \resizebox{\columnwidth}{!}{%
  \begin{tabular}{l c c}
    \toprule
    \textbf{Supervision source} & \textbf{BCB Clone Acc} & \textbf{CC Tag F1} \\
    \midrule
    (1) Human docstrings, as-is                                 & 0.739 $\pm$ 0.010 & 0.229 $\pm$ 0.061 \\
    (2) GPT-4o paraphrase of the human docstring (no code)      & 0.728 $\pm$ 0.006 & 0.230 $\pm$ 0.085 \\
    (3) GPT-4o synthetic, from code                             & \textbf{0.752 $\pm$ 0.003} & \textbf{0.260 $\pm$ 0.016} \\
    (4) Qwen2.5-Coder-7B synthetic, from code                   & 0.749 $\pm$ 0.010 & 0.256 $\pm$ 0.025 \\
    \bottomrule
  \end{tabular}}
  \caption{Controlled 50k supervision-source suite. Identical 50k code subset of the 1S corpus; arms differ only in the supervision text; identical pretraining recipe and fine-tune/eval protocol as Table~\ref{tab:ablations}; 3 seeds.}
  \label{tab:supervision_source}
\end{table}

\section{Description properties}
\label{app:descprops}

\paragraph{\textbf{Lexical and Syntactic properties}} To analyze the differences between synthetic and human docstrings and interpret ablation results, the study employs three metrics: average token length, lexical diversity~\cite{mccarthy2010mtld} (measured via Shannon entropy, the higher the entropy the more diverse the prompt), and identifier overlap \cite{panthaplackel2020associating} (the percentage of code identifiers reused in the docstring, the higher the overlap, the more terms from the code are reused in the prompt). Results are given in Table~\ref{tab:lexical_analysis}. Synthetic descriptions are, on average, three times longer than human docstrings, which also exhibit three times greater length variability, suggesting inconsistent quality in human-written docstrings. Beyond their brevity, human docstrings show lower lexical diversity, meaning they convey less information per token. Additionally, human docstrings have a higher identifier overlap, indicating a tendency to focus on low-level details (e.g., variable names) rather than high-level functionality. In contrast, synthetic descriptions prioritize general code understanding.

\begin{table}[t]
  \centering
  \caption{Lexical properties of human-written and synthetic descriptions. Values are reported as mean $\pm$ standard deviation.}
  \label{tab:lexical_analysis}
  \resizebox{\columnwidth}{!}{%
  \begin{tabular}{l c c c}
    \toprule
    \textbf{Source} &
    \textbf{Avg. Tokens} &
    \textbf{Lexical Diversity} &
    \textbf{Identifier Overlap} \\
    \midrule
    Human Docstrings &
    $13.35 \pm 18.24$ &
    $3.05 \pm 0.98$ &
    $23.36\% \pm 19.91\%$ \\
    Synthetic Descriptions &
    $48.65 \pm 5.66$ &
    $5.54 \pm 0.18$ &
    $5.25\% \pm 5.47\%$ \\
    \bottomrule
  \end{tabular}}
\end{table}

\paragraph{\textbf{Synthetic docstring sensitivity to length}} We then repeated the pre-training and fine-tuning process , generating new synthetic docstrings with different fixed maximum lengths. The goal is to see whether the chosen synthetic docstring length has any meaningful impact on the embedding, and on the performance of the approach. Results are given in Table~\ref{tab:token_sensitivity}. Overall, we observe no monotonic or consistent relationship between description length and downstream performance. While increasing length from 25 to 50 tokens improves tag classification, further increasing to 100 tokens does not yield additional gains and can even degrade performance on some tasks. This indicates that description length alone is not a primary driver of representation quality. Importantly, these results do not imply that longer descriptions are inherently better. When considered alongside the lexical and structural properties of synthetic descriptions analyzed above, this suggests that factors beyond raw length, such as consistency and level of abstraction, are likely to play a role.

\begin{table}[t]
  \centering
  \tiny
  \caption{Sensitivity to description length. Metrics are reported on representative downstream tasks.}
  \label{tab:token_sensitivity}
  \begin{tabular}{l c c}
    \toprule
    \textbf{Variant} &
    \textbf{Clone Acc} &
    \textbf{Tag F1 (CC)} \\
    \midrule
    SyncDesc (25 tokens)  & 0.7143 $\pm$ 0.020 & 0.2410 $\pm$ 0.0164 \\
    SyncDesc (50 tokens)  & 0.7239 $\pm$ 0.016 & \textbf{0.2762 $\pm$ 0.0087} \\
    SyncDesc (100 tokens) & \textbf{0.7248 $\pm$ 0.015} & 0.2505 $\pm$ 0.0299 \\
    \bottomrule
  \end{tabular}
\end{table}

\subsection{Example synthetic descriptions vs.\ human docstrings}
\label{app:descprops_examples}

To make the aggregate lexical metrics in Table~\ref{tab:lexical_analysis} tangible, we show three (code, human docstring, synthetic description) triples drawn from the SyncDesc-1S pretraining corpus, spanning a short, a medium, and a long function across distinct algorithmic categories. In each case the human docstring is terse and reuses code identifiers, whereas the synthetic description is longer, abstracts away the concrete names, and states the functional intent, consistent with the averages in Table~\ref{tab:lexical_analysis} (human $\approx$13 tokens, $\approx$23\% identifier overlap; synthetic $\approx$48 tokens, $\approx$5\% overlap).

\paragraph{\textbf{Example 1: numeric aggregation.}}
\begin{lstlisting}
public double addAndEvaluate(Recommendation<U,I> recommendation){
  double v=metric.evaluate(recommendation);
  if (!ignoreNaN || !Double.isNaN(v)) {
    sum+=v;
    if (!allUsers) {
      numUsers++;
    }
  }
  return v;
}
\end{lstlisting}
\noindent\textit{Human docstring:} ``Adds the recommendation metric to the average and returns the user value.''

\noindent\textit{Synthetic description:} ``This routine calculates a score for a given suggestion and updates cumulative statistics. It first computes the score, and if valid under certain conditions, it adds this score to a running total. Additionally, if not all entities are considered, it updates the count of entities. Finally, it returns the computed score for further use or verification.''

\noindent The human docstring (12 tokens) reuses the identifiers \texttt{recommendation} and \texttt{metric}, whereas the synthetic description (56 tokens) uses no code identifiers, abstracting them to ``suggestion'' and ``score''.

\paragraph{\textbf{Example 2: resource installation and error handling.}}
\begin{lstlisting}
@Override public synchronized Extension install(URL url)
    throws InvalidConfigException {
  Preconditions.checkNotNull(url);
  try {
    File tmpFile=download(url);
    Extension extension=loadFromFile(tmpFile);
    finishInstall(tmpFile,extension);
    return extension;
  } catch (InvalidConfigException e) {
    throw e;
  } catch (Exception e) {
    String msg=baseAction.getText(
        "admin.extension.install.error",
        new String[]{url.toString()});
    log.error(msg,e);
    throw new InvalidConfigException(TYPE.INVALID_EXTENSION,msg,e);
  }
}
\end{lstlisting}
\noindent\textit{Human docstring:} ``Download and install an extension into local file. The final filename is based on the extension's rowType.''

\noindent\textit{Synthetic description:} ``This routine ensures the download and integration of a software component from a network location. It verifies the input's validity, retrieves the component, and validates its configuration. If errors occur, it logs the issue and raises a specific exception. Upon successful validation, it finalizes the integration process of the component and returns it for further operations.''

\noindent The human docstring (17 tokens) names \texttt{extension} and \texttt{rowType}, whereas the synthetic description (56 tokens) summarizes the download/validate/finalize control flow in domain-neutral terms with negligible identifier overlap.

\paragraph{\textbf{Example 3: data-structure manipulation.}}
\begin{lstlisting}
public void reconnectCircuit(){
  ArrayList<Gate> new_Gates=new ArrayList<Gate>();
  ArrayList<Wire> new_Wires=new ArrayList<Wire>();
  if (_Gates != null) {
    for (int i=0; i < _Gates.size(); i++) {
      Gate g=new Gate(_Gates.get(i));
      new_Gates.add(g);
    }
  }
  if (_Wires != null) {
    for (int i=0; i < _Wires.size(); i++) {
      Wire w=new Wire(_Wires.get(i));
      new_Wires.add(w);
    }
  }
  for (int i=0; i < new_Gates.size(); i++) {
    if (_Gates.get(i).Outgoing != null) {
      int index=_Gates.get(i).Outgoing.Index;
      for (Wire w : new_Wires) {
        if (w.Index == index) new_Gates.get(i).Outgoing=w;
      }
    }
  }
  for (int i=0; i < new_Wires.size(); i++) {
    if (_Wires.get(i).From != null) {
      int index=_Wires.get(i).From.Index;
      for (Gate g : new_Gates)
        if (g.Index == index) new_Wires.get(i).From=g;
    }
    if (_Wires.get(i).To != null) {
      int index=_Wires.get(i).To.Index;
      for (Gate g : new_Gates)
        if (g.Index == index) new_Wires.get(i).To=g;
    }
    if (_Wires.get(i).Next != null) {
      int index=_Wires.get(i).Next.Index;
      for (Wire w : new_Wires)
        if (w.Index == index) new_Wires.get(i).Next=w;
    }
  }
  _Gates=new_Gates;
  _Wires=new_Wires;
}
\end{lstlisting}
\noindent\textit{Human docstring:} ``Gate and Wire objects must `point' to each other in memory in a connected circuit.''

\noindent\textit{Synthetic description:} ``This routine reestablishes network connections by duplicating and reassigning components. It replicates nodes and links, then updates relationship references, ensuring each node's outputs and links' endpoints correctly point to the new entities. By iterating over original components, it reconstructs connections in the newly formed structure, maintaining the original network's topology.''

\noindent The human docstring (15 tokens) leans on the concrete types \texttt{Gate} and \texttt{Wire}, whereas the synthetic description (50 tokens) abstracts them to ``nodes'' and ``links'' and summarizes the topology-preserving reconstruction.

\section{Latent-space (UMAP) analysis}
\label{app:umap}

\begin{figure*}[t]
  \centering
  \includegraphics[width=\textwidth]{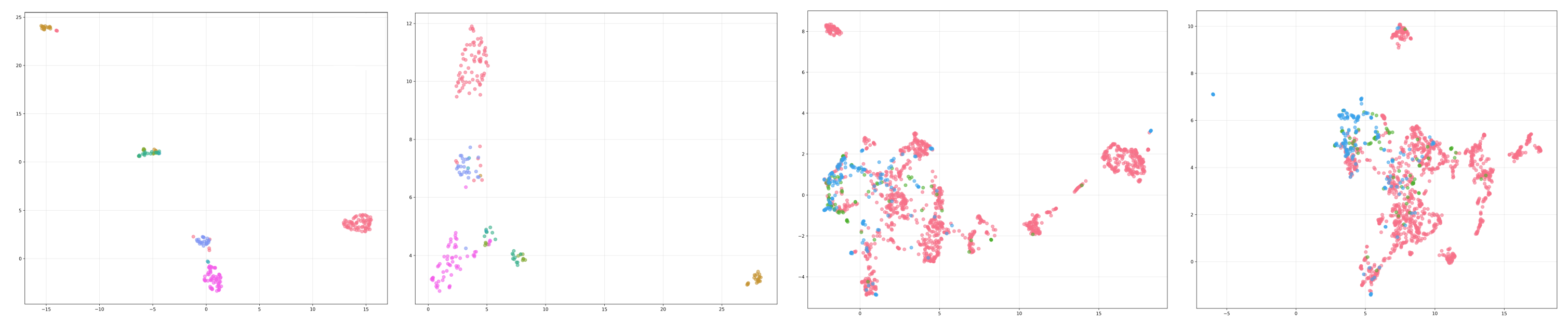}
  \caption{UMAP projections of code embeddings trained with \textit{synthetic descriptions} (first and third panels) and \textit{human docstrings} (second and fourth panels). Colors indicate semantic algorithmic labels (first two panels) and non-semantic difficulty labels (last two panels).}
  \label{fig:umap}
\end{figure*}

To complement the quantitative cluster-separation analysis in Section~\ref{sec:results}, we project the learned embeddings using Uniform Manifold Approximation and Projection (UMAP)~\cite{mcinnes2020umapuniformmanifoldapproximation} for 2D visualization. Semantic alignment is evaluated using algorithmic category tags (e.g., sorting) and difficulty labels from the CodeChef (Java) test set. Figure~\ref{fig:umap} compares UMAP projections of embeddings trained with synthetic descriptions and human-written docstrings. Visually, embeddings from synthetic descriptions exhibit tighter clustering for algorithmic types, suggesting better semantic alignment, while difficulty labels show no clear clustering, likely because difficulty does not directly correlate with semantic content.

\section{Decontamination audit}
\label{app:decontamination}

We audit overlap between both pretraining corpora and every downstream test set. Exact duplicates are detected via MD5 hashes of whitespace-stripped, lowercased code; near-duplicates via MinHash-LSH (128 permutations, 5-gram token shingles, Jaccard $\geq 0.7$). Priority was given to the GitHub-Java channel (FunCom/DeepCom/CONCODE against the CodeSearchNet-Java test set and BigCloneBench), where overlap risk is most plausible since pretraining and evaluation draw on overlapping repository ecosystems. Table~\ref{tab:decontamination} reports the counts.

\begin{table}[t]
  \centering
  \resizebox{\columnwidth}{!}{%
  \begin{tabular}{l l r r}
    \toprule
    \textbf{Pretraining corpus} & \textbf{Test set (items)} & \textbf{Exact} & \textbf{Near ($J \geq 0.7$)} \\
    \midrule
    Java 1S (97{,}252 unique) & Docstring Match (25{,}246)            & 3 (0.01\%)   & 102 (0.40\%) \\
    Java 1S                   & CodeChef Java (1{,}890)               & 0            & 0 \\
    Java 1S                   & BigCloneBench functions (830{,}832)   & 911 (0.11\%) & see text \\
    C / CodeNet (18{,}266 unique) & POJ-104 clone (12{,}000)          & 0            & 0 \\
    C / CodeNet               & CodeForces eval (1{,}370{,}195)       & 0            & n/a \\
    \bottomrule
  \end{tabular}}
  \caption{Decontamination audit: overlap between pretraining corpora and test sets. Exact = MD5 on normalized code; Near = MinHash-LSH near-duplicates.}
  \label{tab:decontamination}
\end{table}

For BigCloneBench, our MinHash configuration proved unreliable on very short functions, so we report the verified exact-hash count (911 functions, 0.11\%; the affected test-\emph{pair} fraction is lower, since a pair is affected only if it contains an overlapping function). Given these magnitudes, the expected metric shift on affected metrics is under 0.5\%; all C-side test sets have zero overlap.

On generator-side leakage: description generation was conditioned only on pretraining-corpus code, and no test items were ever sent to the GPT-4o API, so the synthetic supervision cannot encode test-set content. In preparing these exact statistics, we found that the originally stated ``approximately 200k Java training samples'' described the aggregated description pool rather than the pretrained subset; the pretraining sets contain exactly 99{,}991 Java samples (97{,}252 unique implementations) and 24{,}936 C samples (18{,}266 unique), as now stated in Section~\ref{sec:experimental_setup-datasets}. This strengthens rather than weakens our comparisons: the same results are obtained from roughly half the previously stated data volume, and the data-volume gap to ContraCode widens to $\sim$15$\times$.

\section{Description generation: prompt, statistics, and cost}
\label{app:generation}

\paragraph{\textbf{Prompt and decoding.}} Descriptions are generated with GPT-4o (temperature 0.7, top-$p$ 1.0, maximum 70 generated tokens, with the prompt instructing at most 50 tokens; Section~\ref{sec:experimental_setup}). The prompt wrapper adds 180 tokens per call. The exact prompt and decoding scripts are provided verbatim in the replication package.

\paragraph{\textbf{Filtering and failure statistics.}} The aggregated Java description pool comprised $\sim$200k generated code--description pairs; after filtering and generation failures, the pretraining set contains 99{,}991 samples (97{,}252 unique implementations). The C corpus contains 24{,}936 samples (18{,}266 unique implementations). The complete generation logs, with per-corpus filtering and failure counts, are shipped with the replication package.

\paragraph{\textbf{One-time supervision-generation cost.}} Generation ran via the GPT-4o Batch API (USD 1.25 per million input tokens, USD 5 per million output tokens; rates verified 2026-07-11, unchanged since August 2024 and therefore applicable at generation time). Table~\ref{tab:generation_cost} gives the full accounting: USD 91 of one-time generation in total, followed by 8 A100-hours of contrastive pretraining. Per-task fine-tuning cost is one epoch on a data subset, and the pretrained encoder is reused across all tasks. As shown in Table~\ref{tab:supervision_source}, an open-weights generator (Qwen2.5-Coder-7B) matches GPT-4o within seed noise, so the generation cost can drop to essentially zero on a local GPU.

\begin{table}[t]
  \centering
  \resizebox{\columnwidth}{!}{%
  \begin{tabular}{l r r r r}
    \toprule
    \textbf{Scope} & \textbf{Samples} & \textbf{Input tok.} & \textbf{Output tok.} & \textbf{Cost (Batch API)} \\
    \midrule
    1S Java corpus      & 99{,}991  & 24.3M & 7.82M & USD 70 \\
    C corpus (CodeNet)  & 24{,}936  & 9.4M  & 1.96M & USD 22 \\
    \midrule
    2S total            & 124{,}927 & 33.7M & 9.78M & USD 91 \\
    \bottomrule
  \end{tabular}}
  \caption{One-time supervision-generation cost, as run.}
  \label{tab:generation_cost}
\end{table}

\paragraph{\textbf{Comparison with execution-trace collection (TRACED).}} TRACED's analogous one-time cost is execution infrastructure: compilable programs paired with valid inputs, per-sample multi-input runs, and a language-specific gcc/gdb tracing toolchain. Rerunning their collection pipeline on the same CodeNet C corpus required about 4{,}020 sequential CPU-hours; the tracer, not compilation, is the bottleneck, since gdb single-steps each program with a per-line variable snapshot. This supervision is also language-bound (the toolchain must be rebuilt per language), whereas description generation applies uniformly across languages. Zero-shot LLM usage, by contrast, incurs API calls for every task and query, and raises privacy issues for private code.

\paragraph{\textbf{Inference throughput.}} After pretraining, inference runs at roughly 204 snippets/s even unbatched on a single consumer GPU (RTX 4090, fp16, sequence length 512): embedding the 5{,}000 POJ-104 test snippets takes 24.5\,s under the identical embedding pipeline used for Table~\ref{tab:scale_comparison}, one forward pass per snippet (a conservative lower bound of over 17M snippets/day), with no per-query API cost, latency, or rate limits. It also means no code egress, which matters for private codebases where per-query API calls are often prohibited but a one-time generation pass over a vetted corpus is permissible.

\end{document}